\documentclass[conference]{IEEEtran}
\IEEEoverridecommandlockouts

\usepackage{amsmath,amssymb,amsfonts}
\usepackage{algorithm}
\usepackage{algpseudocode}
\usepackage{graphicx}
\usepackage{textcomp}
\usepackage{xcolor}
\usepackage{booktabs}
\usepackage{multirow}
\usepackage{ulem}
\usepackage{subcaption}
\usepackage[table]{xcolor}
\usepackage{float}

\def\BibTeX{{\rm B\kern-.05em{\sc i\kern-.025em b}\kern-.08em
    T\kern-.1667em\lower.7ex\hbox{E}\kern-.125emX}}

\usepackage[draft,bookmarks=false]{hyperref}

\begin{document}

\title{Efficient and Adaptive Human Activity Recognition via LLM Backbones}

\author{\IEEEauthorblockN{ Aleksandr BREDIKHIN}
\IEEEauthorblockA{Univ. Grenoble Alpes, France \\
aleksandr.bredikhin@etu.univ-grenoble-alpes.fr}
\and

\IEEEauthorblockN{Philippe LALANDA}
\IEEEauthorblockA{Univ. Grenoble Alpes, France \\
philippe.lalanda@imag.fr}
\and
\IEEEauthorblockN{German VEGA}
\IEEEauthorblockA{Univ. Grenoble Alpes, France \\
german.vega@imag.fr}

}

\maketitle

\begin{abstract}
Human Activity Recognition (HAR) is a core task in pervasive computing systems, where models must operate under strict computational constraints while remaining robust to heterogeneous and evolving deployment conditions. Recent advances based on Transformer architectures have significantly improved recognition performance, but typically rely on task-specific models trained from scratch, resulting in high training cost, large data requirements, and limited adaptability to domain shifts. In this paper, we propose a paradigm shift that reuses large pretrained language models (LLMs) as generic temporal backbones for sensor-based HAR, instead of designing domain-specific Transformers. To bridge the modality gap between inertial time series and language models, we introduce a structured convolutional projection that maps multivariate accelerometer and gyroscope signals into the latent space of the LLM. The pretrained backbone is kept frozen and adapted using parameter-efficient Low-Rank Adaptation (LoRA), drastically reducing the number of trainable parameters and the overall training cost. Through extensive experiments on standard HAR benchmarks, we show that this approach enables rapid convergence, strong data efficiency, and robust cross-dataset transfer, particularly in low-data and few-shot settings. At the same time, our results highlight the complementary roles of convolutional frontends and LLMs, where local invariances are handled at the signal level while long-range temporal dependencies are captured by the pretrained backbone. Overall, this work demonstrates that LLMs can serve as a practical, frugal, and scalable foundation for adaptive HAR systems, opening new directions for reusing foundation models beyond their original language domain.
\end{abstract}

\maketitle

\section{Introduction}

Pervasive computing systems increasingly rely on machine learning (ML) to interpret large volumes of multimodal sensor data and enable intelligent behavior in applications such as wearable health monitoring, activity tracking, smart environments, and context-aware mobile services. In these settings, models must operate continuously under strict computational, energy, and data constraints while maintaining reliable performance across heterogeneous and evolving operating conditions. Achieving both efficiency and robustness therefore remains a central challenge for pervasive and IoT systems.

Human Activity Recognition (HAR) is a representative and long-standing problem in this domain. HAR models are typically trained on time-series data collected from inertial and physiological sensors and are expected to generalize across users, devices, sensor placements, and environmental contexts. However, real-world deployments routinely expose these models to distributional changes that were not observed during training. 

Recent advances in HAR have largely been driven by deep sequence models, most notably Transformer-based architectures. 
Despite their effectiveness, most Transformer-based approaches are designed and trained from scratch for the HAR domain, requiring large amounts of labeled data, substantial computational resources, and careful architectural tuning. These requirements limit their practicality in pervasive settings, where rapid adaptation to the target  is essential.

At the same time, the challenge of adapting HAR models to heterogeneous domain shifts remains only partially addressed. Existing solutions often rely on domain-specific fine-tuning, explicit alignment objectives, or invariant representation learning. While effective in controlled scenarios, these approaches frequently introduce additional training complexity and remain sensitive to the availability of labeled data in the target domain. Moreover, they implicitly assume static training conditions and struggle to scale efficiently when new domains or devices are introduced after deployment.

In this work, we explore an alternative direction based on the reuse of large pretrained models. We propose to repurpose Large Language Models (LLMs) as generic temporal backbones for sensor-based HAR, moving away from the conventional paradigm of designing task-specific Transformers from scratch. Although originally developed for natural language processing, LLMs encode rich sequence representations learned through large-scale pretraining. Our key insight is that the temporal modeling capabilities learned from linguistic data can be transferred to inertial time series, provided that an appropriate modality alignment is performed. Rather than relying on prompting-based strategies, which are ill-suited for continuous multivariate signals, we adopt an alignment-based approach that maps sensor data directly into the latent space of pretrained LLMs.

Specifically, we introduce a \textit{modality-aligned architecture} that combines a lightweight convolutional frontend with a pretrained LLM backbone. The convolutional component is responsible for extracting structured and locally invariant representations from raw accelerometer and gyroscope signals, while the LLM operates as a high-level temporal reasoner capturing long-range dependencies and activity transitions. The LLM backbone is kept frozen or lightly adapted using parameter-efficient Low-Rank Adaptation (LoRA), drastically reducing the number of trainable parameters and the overall training cost while preserving  temporal modeling capabilities.

Beyond efficiency, this design provides favorable properties for adaptation. The pretrained latent space of LLMs offers a strong representational prior that supports rapid convergence, few-shot fine-tuning, and robust cross-dataset transfer. At the same time, our results indicate that certain forms of variability, such as unconstrained sensor placement, must be addressed at the convolutional level, highlighting the complementary roles of domain-specific inductive bias and large-scale pretraining.

Through extensive experiments on standard HAR benchmarks, we demonstrate that the proposed approach achieves competitive or state-of-the-art performance while requiring significantly fewer training resources than Transformer-based baselines. Strong results are obtained in low-data and cross-dataset scenarios, underscoring the relevance of LLM backbones for realistic pervasive deployments where labeled data is scarce and operating conditions evolve over time.

The main contributions of this paper are as follows:
\begin{itemize}
    \item We introduce a paradigm shift for HAR by \textbf{reusing pretrained LLMs as temporal backbones}, rather than designing task-specific Transformer architectures from scratch.
    \item We propose a \textbf{convolutional projection and LoRA-based adaptation framework} that enables efficient alignment between inertial time series and LLM latent spaces.
    \item We analyze the \textbf{complementary roles of convolutional encoders and LLMs}, highlighting their respective contributions to local invariance and long-range temporal modeling.
    \item We demonstrate that the proposed approach achieves \textbf{strong performance, data efficiency, and transferability} under realistic resource and supervision constraints.
\end{itemize}

The remainder of this paper is organized as follows. Section~2 reviews related work on sensor-based HAR, Transformer architectures, and LLM-based time-series modeling. Section~3 presents the proposed architecture and alignment strategy. Section~4 details the implementation and training protocol. Section~5 reports experimental results on multiple HAR benchmarks. Section~6 discusses the implications and limitations of the approach. Section~7 concludes the paper and outlines directions for future work.

\section{State of the Art}

\subsection{Sensor-based Human Activity Recognition}

Human Activity Recognition (HAR) aims to automatically identify human actions from data capturing body movements and activity patterns, and constitutes a key enabling technology for healthcare, assisted living, and security applications. While alternative sensing modalities such as video- and WiFi-based approaches have been explored~\cite{onofri2016survey}, they often raise privacy concerns and require fixed infrastructure. In contrast, sensor-based HAR relies on Inertial Measurement Units (IMUs) embedded in wearable and mobile devices, offering a privacy-preserving, low-cost, and scalable solution, which is the focus of this work.

Among available sensing modalities, accelerometers and gyroscopes have consistently proven the most informative for activity recognition~\cite{wang2019deep,ignatov2018real}. Accelerometers capture linear motion intensity, while gyroscopes encode angular velocity and orientation, providing complementary information that is critical for distinguishing activities with similar acceleration patterns but different motion dynamics. These two modalities form the standard multivariate time-series representation adopted in this study.

\subsection{Deep Learning for HAR}

Deep learning has become the dominant paradigm for HAR, progressively replacing handcrafted feature pipelines. Convolutional Neural Networks (CNNs) enabled hierarchical feature extraction directly from raw sensor signals~\cite{ordonez2016deepconv,hammerla2016deep}, while Recurrent Neural Networks (RNNs), and in particular Long Short-Term Memory networks (LSTMs)~\cite{hochreiter1997long}, were introduced to capture longer temporal dependencies. Hybrid architectures such as DeepConvLSTM emerged as strong baselines, but remain constrained by limited receptive fields, sequential processing bottlenecks, and reduced robustness under heterogeneous deployment conditions.

Transformer architectures, originally introduced for natural language processing~\cite{vaswani2017attention}, address several of these limitations through self-attention mechanisms that enable global temporal modeling and parallel computation. Transformer-based models have achieved state-of-the-art performance on multiple HAR benchmarks~\cite{dirgova2021transformer,IFConvTransformer2022}. More recent architectures such as HART  further incorporate domain-specific inductive biases to improve robustness across devices and sensor placements~\cite{ek2022hart}. Despite their effectiveness, these models are typically trained from scratch on HAR data, requiring substantial labeled datasets and significant computational resources. Such requirements limit their practicality in pervasive settings, where labeled data is scarce and rapid adaptation to new conditions is often required \cite{ek2024comparing}.

\subsection{Large Language Models for Time Series Analysis}

Large Language Models (LLMs) have demonstrated remarkable capabilities in modeling long-range dependencies through large-scale pretraining, and recent work has shown that they can be adapted beyond text to numerical and temporal domains. A recent survey~\cite{zhang2024largelanguagemodelstime} categorizes these approaches into prompting-based, aligning-based, and hybrid strategies.

Prompting-based methods treat time series as serialized text and directly query pretrained LLMs. While attractive in terms of simplicity, such approaches suffer from fundamental limitations when applied to sensor data, including poor numerical precision, limited context length, and inadequate modeling of fine-grained temporal dynamics. As a result, their applicability to tasks such as HAR remains limited.

Aligning-based approaches offer a more principled alternative by explicitly mapping time-series data into the latent space of pretrained LLMs, either through learned projection layers or contrastive alignment objectives. In this paradigm, LLMs are reused as backbone sequence encoders, while modality-specific components handle the adaptation from continuous signals to token-level representations. This strategy preserves the benefits of large-scale pretraining while explicitly addressing the modality mismatch between language and sensor data.

The increasing availability of open pretrained models—including GPT-style architectures~\cite{radford2019language}, the LLaMA family~\cite{touvron2023llama}, and multilingual LLMs such as Qwen~\cite{bai2023qwen}—has made LLMs practical building blocks for downstream applications where training large models from scratch is infeasible. However, despite growing interest in LLMs for time-series analysis, their use as backbone architectures for wearable sensor-based HAR remains largely unexplored. To the best of our knowledge, no prior work has systematically investigated aligning-based LLM architectures for inertial HAR, leaving an open gap that this work aims to address.

The adoption of LLMs in downstream tasks has motivated the development of Parameter-Efficient Fine-Tuning (PEFT) methods, which aim to reduce the cost and instability associated with full fine-tuning. Existing approaches include adapter-based modules~\cite{houlsby2019parameter}, prefix- and prompt-tuning techniques~\cite{li2021prefix,lester2021power}, and low-rank decomposition methods. Among these, Low-Rank Adaptation (LoRA)~\cite{hu2021lora} has gained particular attention due to its simplicity and effectiveness, injecting trainable low-rank matrices into transformer layers while keeping the original weights frozen.

\section{Proposed Approach}

In contrast to existing Transformer-based HAR models that rely on task-specific architectures trained from scratch, we propose to reuse Large Language Models (LLMs) as pretrained sequence backbones for sensor-based Human Activity Recognition. While originally developed for natural language processing, LLMs have demonstrated strong capabilities in modeling complex sequential dependencies and organizing information into highly structured latent spaces. Importantly, these capabilities are not inherently tied to language, but to sequence modeling itself. However, directly applying language modeling techniques to raw inertial time series is ineffective due to the fundamental mismatch between discrete linguistic tokens and continuous, multivariate sensor signals.

Our key insight is to treat LLMs as \textbf{generic temporal encoders} rather than language-specific models, and to bridge the modality gap through an explicit alignment mechanism. Instead of redesigning Transformer architectures for each sensing domain, we introduce a lightweight, task-specific frontend that maps inertial signals into a representation compatible with pretrained LLMs. This strategy allows us to maximally reuse existing foundation models while retaining the inductive biases required for sensor-based HAR.

\begin{figure*}[t]
    \centering
    \begin{subfigure}{0.9\textwidth}
        \centering
        \includegraphics[width=\linewidth]{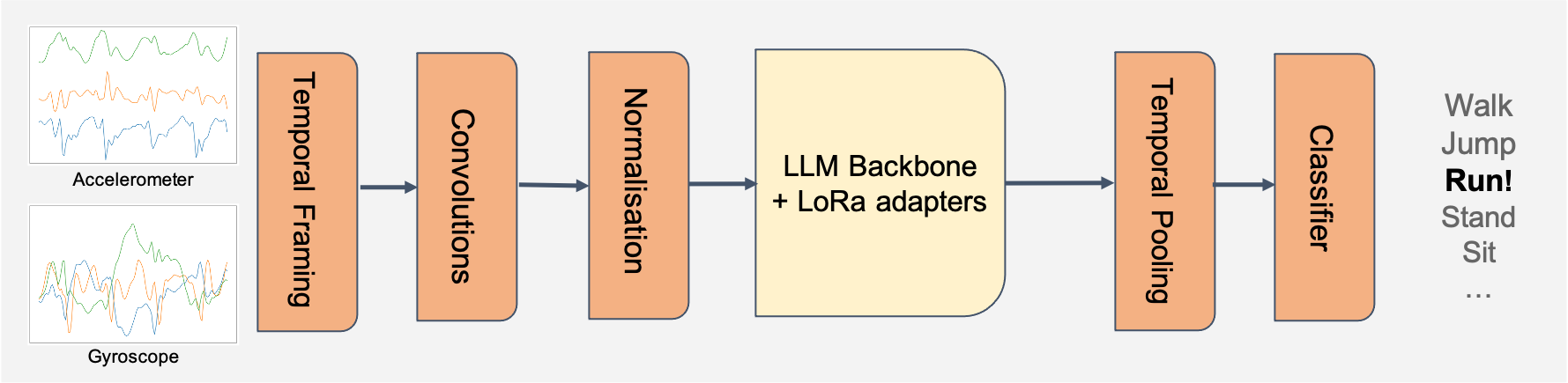}
        \caption{Overall architecture of the proposed LLM-based HAR framework.}
        \label{fig:archi_llm}
    \end{subfigure}

    \vspace{0.5em}

    \begin{subfigure}{0.6\textwidth}
        \centering
        \includegraphics[width=\linewidth]{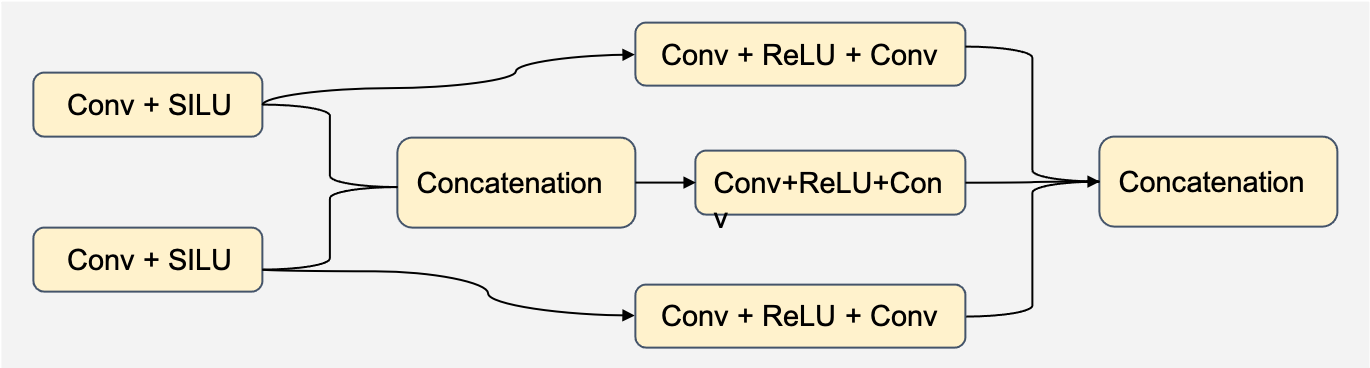}
        \caption{Convolutional projection module mapping inertial signals to LLM embeddings.}
        \label{fig:conv_proj}
    \end{subfigure}

    \caption{Overview of the proposed architecture. (a) End-to-end pipeline from inertial signals to activity classification. (b) Convolutional projection used to adapt multivariate inertial time series to the LLM input space.}
    \label{fig:architecture}
\end{figure*}

The proposed framework, called HARLLM and illustrated in Fig.~\ref{fig:architecture}, is built around three core design principles:
\begin{enumerate}
    \item \textbf{Modality-aligned projection}: a convolutional frontend transforms multivariate accelerometer and gyroscope signals into structured embeddings that preserve local temporal dynamics while being compatible with LLM token representations;
    \item \textbf{Frozen backbone with parameter-efficient adaptation}: the pretrained LLM is kept frozen and adapted via Low-Rank Adaptation (LoRA), enabling efficient specialization with a minimal number of trainable parameters;
    \item \textbf{Decoupled representation and reasoning}: low-level signal processing and invariance handling are delegated to the convolutional module, while the LLM focuses on modeling long-range temporal dependencies and activity-level structure.
\end{enumerate}

This design explicitly separates concerns between signal-level representation and sequence-level reasoning. The convolutional projection module handles sensor-specific properties such as sampling characteristics, noise, and local temporal patterns, and is responsible for enforcing inductive biases related to inertial sensing. The LLM backbone, operating on the resulting window-level embeddings, captures long-range temporal dependencies, activity transitions, and global sequence structure. Crucially, this separation allows each component to operate in the regime where it is most effective, without modifying the architecture of the pretrained LLM.

This architectural choice is particularly well suited to HAR for several reasons:
\begin{itemize}
    \item \textit{Long-range temporal modeling}: pretrained LLMs excel at capturing dependencies over long sequences, which naturally arise in activity streams;
    \item \textit{Data efficiency}: the pretrained latent space provides a strong initialization, enabling rapid convergence and effective learning under limited supervision;
    \item \textit{Domain adaptability}: the structured representations learned by LLMs facilitate transfer across users, devices, and sensing conditions;
    \item \textit{Computational frugality}: freezing the backbone and relying on LoRA significantly reduces training time, memory footprint, and energy consumption.
\end{itemize}

To enable efficient adaptation, we employ LoRA to inject a small number of trainable parameters into selected attention layers of the LLM, while keeping the remainder of the backbone frozen. This strategy preserves the expressive power of the pretrained model while drastically reducing the cost of training and fine-tuning, making the approach practical for rapid experimentation and deployment.

Beyond efficiency, the proposed architecture is designed to handle deployment-time domain shifts. Real-world HAR systems inevitably face variability in users, devices, and sensor placement. By combining a strong pretrained temporal backbone with a lightweight and modular frontend, our framework supports fast specialization to new domains with limited labeled data. At the same time, our experiments indicate that certain forms of variability, such as unconstrained sensor orientation, must be addressed at the convolutional level, further emphasizing the complementary roles of the two components.

Overall, the proposed approach reframes the design of HAR models around the reuse of pretrained foundation models. By explicitly aligning inertial time series with LLM latent spaces and relying on parameter-efficient adaptation, we address three persistent challenges in HAR: high training cost, limited robustness to domain shifts, and data scarcity. This paradigm enables a favorable trade-off between performance, efficiency, and adaptability, and provides a general blueprint for integrating LLMs into pervasive time-series modeling pipelines.

\section{Implementation}

This section details the concrete instantiation of the proposed LLM-based HAR framework. We describe the input representation, the convolutional projection module used to adapt inertial time series to the LLM latent space, the pretrained backbone with parameter-efficient adaptation, and the final classification pipeline. An overview of the end-to-end architecture is shown in Fig.~\ref{fig:archi_llm}, while the internal structure of the convolutional projection module is illustrated in Fig.~\ref{fig:conv_proj}.

\subsection{Input Representation}

The input consists of multivariate inertial time series collected from wearable devices. Following standard practice in sensor-based HAR, we use tri-axial accelerometer and gyroscope signals. Each input sample is represented as a fixed-length temporal window:
\[
X \in \mathbb{R}^{B \times W \times C},
\]
where $B$ denotes the batch size, $W = 128$ is the window length, and $C = 6$ corresponds to the three accelerometer axes and three gyroscope axes. Each channel is independently $z$-normalized to reduce inter-sensor scale variability. Temporal windows are extracted using an overlapping sliding-window strategy.

\subsection{Temporal Framing}

Since LLMs operate on sequences of discrete tokens, continuous time series must be converted into a sequence of latent frames. We adopt a framing strategy that segments each input window into shorter, non-overlapping temporal segments. Given a frame length $L = 16$, each window is decomposed into
\[
N = \frac{W}{L} = 8
\]
frames, yielding the framed representation:
\[
X_f \in \mathbb{R}^{B \times N \times L \times C}.
\]
Each frame captures a short temporal context and will be mapped to a single latent token for the LLM.

\subsection{Convolutional Projection Module}

To bridge the modality gap between continuous inertial signals and discrete LLM tokens, we introduce a convolutional projection module that maps each temporal frame to a structured embedding. All convolutional operations are one-dimensional (1D) and operate along the temporal dimension of each frame. This module serves as a \textbf{modality adapter}, extracting locally invariant motion patterns while producing embeddings compatible with the LLM input space.

\paragraph{Modality-specific encoding.}
Accelerometer and gyroscope signals are processed independently to preserve their distinct physical characteristics. For each modality $m \in \{\text{acc}, \text{gyro}\}$, we apply a temporal convolution within each frame:
\[
H^{(m)}_0 = \sigma\!\left(\mathrm{Conv1D}^{(m)}(X^{(m)}_f)\right),
\]
where $\mathrm{Conv1D}^{(m)}$ operates along the temporal axis, and $\sigma(\cdot)$ denotes the SiLU activation. This step extracts low-level motion features specific to each modality.

\paragraph{Feature fusion.}
To explicitly combine linear acceleration and rotational dynamics, modality-specific features are concatenated along the channel dimension:
\[
H^{(\text{union})}_0 = \bigl[ H^{(\text{acc})}_0 \,;\, H^{(\text{gyro})}_0 \bigr].
\]

\paragraph{Multi-branch temporal processing.}
To capture complementary temporal dynamics at different abstraction levels, each of the three representations (accelerometer, gyroscope, and fused union) is processed independently through a two-layer temporal convolutional stack. For each branch $b \in \{\text{acc}, \text{gyro}, \text{union}\}$, we compute:
\[
H^{(b)}_1 = \mathrm{Conv1D}^{(b)}_2\!\left(\mathrm{ReLU}\!\left(\mathrm{Conv1D}^{(b)}_1(H^{(b)}_0)\right)\right).
\]
This hierarchical design progressively increases the receptive field, enabling the extraction of both short-term and mid-range temporal patterns within each frame.

\paragraph{Projection to the LLM embedding space.}
Each branch output is linearly projected to a subspace of dimension $d_{\text{LLM}}/3$, where $d_{\text{LLM}}$ denotes the embedding size of the LLM backbone (e.g., $d_{\text{LLM}}=768$ for GPT-2 Small). Let $P^{(b)}$ denote the projection matrix for branch $b$. We obtain:
\[
\tilde{H}^{(b)}_1 = H^{(b)}_1 P^{(b)}.
\]
The projected representations are then concatenated to form a unified embedding:
\[
H_2 = \bigl[ \tilde{H}^{(\text{acc})}_1 \,;\, \tilde{H}^{(\text{gyro})}_1 \,;\, \tilde{H}^{(\text{union})}_1 \bigr] \in \mathbb{R}^{B \times N \times d_{\text{LLM}}}.
\]
Finally, Layer Normalization is applied to ensure scale consistency and stable integration with the pretrained language model:
\[
E = \mathrm{LN}(H_2).
\]
The resulting token sequence $E$ preserves temporal ordering while encoding local motion semantics suitable for LLM processing.

\subsection{LLM Backbone and Parameter-Efficient Adaptation}

The normalized token sequence $E$ is fed into a pretrained LLM backbone, which acts as a generic sequence encoder. In our main experiments, we use GPT-2 Small as backbone, comprising approximately 124M parameters. This choice provides a favorable trade-off between representational capacity and computational efficiency, allowing us to remain within modest model sizes while preserving strong temporal modeling capabilities. To assess the robustness of the proposed framework with respect to the choice of backbone, we additionally experimented with other pretrained LLMs of comparable or moderately larger scale, including GPT-2 Medium (approximately 355M parameters) and Qwen-1.8B. Despite differences in architecture and pretraining data, these models exhibited similar performance trends and comparable recognition accuracy once integrated into our alignment-based framework. 


In all experiments, the LLM backbone parameters are kept frozen during training to preserve pretrained knowledge and limit computational cost. Adaptation is performed exclusively through LoRA modules inserted into the query, key, and value projections of the self-attention layers. For each adapted projection matrix $W \in \mathbb{R}^{d \times d}$, LoRA introduces a low-rank update of the form
\[
W' = W + BA,
\]
where $A \in \mathbb{R}^{r \times d}$ and $B \in \mathbb{R}^{d \times r}$ are trainable matrices, and $r$ denotes the adaptation rank.

In our implementation, we set $r=16$ and apply LoRA to all self-attention layers of the backbone. This results in approximately $1.2$ million trainable parameters for GPT-2 Small, corresponding to less than $1\%$ of the total backbone size. The LoRA scaling factor is set to $\alpha=32$, and a dropout rate of $0.05$ is applied within the LoRA modules for regularization. This configuration provides sufficient adaptation capacity while maintaining strong regularization in low-data regimes.

Compared to training a Transformer-based HAR model from scratch, which typically requires updating tens of millions of parameters and extensive task-specific training, the proposed approach drastically reduces both the number of trainable parameters and the overall training cost. In practice, this enables rapid fine-tuning, lower memory usage, and significantly shorter training times.

\subsection{Temporal Pooling and Classification}

The LLM outputs a sequence of contextualized embeddings $\{h_t\}_{t=1}^{N}$. To obtain a fixed-dimensional representation for activity classification, we apply temporal mean pooling:
\[
h = \frac{1}{N} \sum_{t=1}^{N} h_t.
\]
The pooled representation is passed to a lightweight classification head consisting of a linear layer followed by a softmax. The model is trained end-to-end using cross-entropy loss, updating only the convolutional projection module, the LoRA parameters, and the classification head.

\section{Experiments}

\subsection{Introduction}

This section evaluates the proposed HARLLM framework along three complementary objectives:
(i) absolute recognition performance,
(ii) data efficiency when training from scratch with limited labeled data, and
(iii) robustness to cross-dataset transfer under constrained supervision.

To cover these objectives in a representative and controlled manner, we conduct experiments on three widely used human activity recognition benchmarks: \textit{HHAR}, \textit{UCI HAR}, and \textit{RealWorld}. Together, these datasets capture the main sources of variability encountered in pervasive sensing, including device heterogeneity, sensor placement, and real-world acquisition conditions \cite{10207558}.

The \textit{HHAR} dataset~\cite{stisen2015smart} serves as our primary benchmark for studying data efficiency and transfer learning, as it includes heterogeneous devices (smartphones and smartwatches), multiple users, and diverse sensor characteristics. The \textit{UCI HAR} dataset~\cite{Anguita2013APD} represents a controlled acquisition setting with fixed sensor placement and well-defined activity patterns, and is used to assess absolute recognition performance under stable conditions. The \textit{RealWorld} dataset~\cite{realword} captures unconstrained real-world scenarios with significant variability in sensor placement and user behavior, making it a challenging benchmark for evaluating robustness under realistic deployment conditions.

Unless otherwise stated, absolute recognition performance is reported on all three datasets, while data efficiency and cross-dataset transfer experiments are conducted using HHAR as the source dataset.

\subsection{Training Protocol and Metrics}

All models are trained with a batch size of 128 using the Adam optimizer, with an initial learning rate of $1.0 \times 10^{-4}$. To ensure stable convergence, early stopping is applied based on the weighted F1-score measured on a validation set. 

We report both weighted F1-score and per-sample accuracy as evaluation metrics. The weighted F1-score accounts for class imbalance by weighting each class according to its support, while per-sample accuracy provides a complementary measure of overall recognition performance.

\subsection{Performance}

We first evaluate recognition performance on the three datasets under standard fully supervised training. HARLLM is compared against representative baselines spanning classical convolutional architectures, hybrid CNN--RNN models, and recent Transformer-based approaches for human activity recognition. Specifically, we include a lightweight CNN baseline~\cite{Ignatov2018RealtimeHA}, the DeepConvLSTM model~\cite{ordonez2016deep} as a strong recurrent baseline, a conventional Transformer adapted to time-series inputs~\cite{dosovitskiy2021an}, as well as HART~\cite{ek2022hart}, which represents a state-of-the-art Transformer-based architecture specifically designed for HAR. Table~\ref{tab:multi_dataset_results} reports weighted F1-scores on the UCI HAR, HHAR, and RealWorld test sets.

\begin{table}
\caption{F1-score with known situations on the UCI, HHAR, and RealWorld test sets.
\textbf{Bold} indicates best results while \underline{underline} indicates second-best results.}
\label{tab:multi_dataset_results}
\centering
\begin{tabular}{l c c c}
\toprule
Architecture & UCI & HHAR & RealWorld \\
\midrule
CNN \cite{Ignatov2018RealtimeHA} 
& 94.53 & 96.91 & 92.62 \\
DeepConvLSTM \cite{ordonez2016deep} 
& \textbf{95.94} & \underline{97.94} & \underline{94.39} \\
Conventional Transformer \cite{dosovitskiy2021an} 
& 93.66 & 96.05 & 93.80 \\
HART \cite{ek2022hart} 
& 94.49 & 97.36 & \textbf{94.88} \\
\midrule
HARLLM [Ours] 
& \underline{95.78} & \textbf{98.43} & 94.16 \\
\bottomrule
\end{tabular}
\end{table}

HARLLM achieves the highest weighted F1-score on HHAR, outperforming both classical convolutional models and recent Transformer-based architectures. On UCI HAR, it reaches competitive performance close to the best-performing baseline, while on RealWorld it remains slightly below HART, which explicitly encodes sensor-placement invariances. Overall, these results indicate that large language models, when properly adapted to time-series inputs, can achieve state-of-the-art or near state-of-the-art performance across heterogeneous HAR benchmarks, while relying on parameter-efficient adaptation rather than task-specific Transformer architectures trained from scratch.

\subsection{Robustness to Training Data Size}

We next analyze data efficiency by training HARLLM from scratch on HHAR using progressively smaller amounts of labeled training data. Results are summarized in Table~\ref{tab:hhar_train_from_srach}.

\begin{table}[t]
\centering
\caption{Results on HHAR when training from scratch with different amounts of labeled data.}
\label{tab:hhar_train_from_srach}
\footnotesize
\begin{tabular}{lcc}
\toprule
Train/Test split & F1-weighted & Accuracy \\
\midrule
80/20    & 98.36 & 98.37 \\
20/80    & 95.46 & 95.49 \\
10/90    & 93.52 & 93.57 \\
1/99     & 73.81 & 74.36 \\
\bottomrule
\end{tabular}
\end{table}

HARLLM maintains strong performance even when trained with substantially reduced amounts of labeled data. Notably, using only 20\% of the training set already yields a weighted F1-score above 95, and competitive performance is preserved down to 10\% of labeled data. This rapid performance saturation suggests that the pretrained LLM backbone provides robust temporal representations that can be effectively adapted with limited supervision. Performance degradation becomes significant only in the extreme low-data regime (1\% of labeled data), where task-specific information is insufficient to fully specialize the model. Overall, these results highlight the data efficiency of HARLLM and support the use of pretrained LLM representations for HAR in scenarios where labeled data is scarce.

\subsection{Cross-Dataset Transfer: Leave-One-Dataset-Out}

To evaluate robustness to domain shifts, we adopt a Leave-One-Dataset-Out (LODO) protocol. In this setting, the model is first pretrained on multiple source datasets---MotionSense, UCI HAR, PAMAP2, and RealWorld---while completely excluding HHAR from the pretraining phase. The pretrained model is then transferred to HHAR and fine-tuned using only a small fraction of labeled target data ($1\%$ or $10\%$). Performance is reported in Table~\ref{tab:lodo_results}.

\begin{table}[t]
\centering
\caption{Performance under the Leave-One-Dataset-Out (LODO) scenario.}
\label{tab:lodo_results}
\begin{tabular}{lccc}
\toprule
Stage & Target Dataset & Training Data & Weighted F1 (\%) \\
\midrule
Pretraining & -- & -- & 87.67 \\
Fine-tuning & HHAR & 1\% & 80.89 \\
Fine-tuning & HHAR & 10\% & 93.68 \\
Fine-tuning & HHAR & 20\% & 94.53 \\
Fine-tuning & HHAR & 50\% & 96.86 \\
Fine-tuning & HHAR & 80\% & 97.43 \\
\bottomrule
\end{tabular}
\end{table}

Table~\ref{tab:lodo_results} reports performance under the Leave-One-Dataset-Out (LODO) protocol, where the model is pretrained on multiple source datasets and transferred to HHAR with varying amounts of labeled target data. Even without seeing any HHAR samples during training, the pretrained model achieves a weighted F1-score of 87.67, indicating that the representations learned during pretraining already capture generic activity-level structure that transfers across datasets.

When fine-tuned with a very small amount of labeled target data, performance improves substantially. With only $1\%$ of HHAR data, HARLLM maintains a weighted F1-score above 80, highlighting the effectiveness of few-shot fine-tuning under severe data scarcity. Increasing supervision to $10\%$ yields a rapid performance jump to 93.68, demonstrating that the pretrained backbone can be efficiently specialized with limited labeled data. As the amount of target data increases further, performance continues to improve but with diminishing returns. At $20\%$ and $50\%$ of labeled data, HARLLM reaches weighted F1-scores of 94.53 and 96.86, respectively, and approaches the fully supervised regime at $80\%$ of labeled data with a score of 97.43. This progressive trend illustrates a smooth transition from transfer-dominated learning to task-specific specialization.

An interesting behavior emerges across data regimes. Fine-tuning consistently provides clear advantages in low-data settings, while gains become marginal as more labeled data becomes available. A plausible explanation is that pretraining introduces a strong inductive bias that acts as an effective form of regularization under data scarcity, but may become less beneficial when sufficient data is available to learn highly task- and subject-specific representations from scratch. This effect is particularly relevant in HAR, where inter-subject variability is high and abundant data enables models to capture fine-grained individual motion patterns. Overall, these results confirm the effectiveness of LLM pretraining for data-efficient cross-dataset transfer, while clarifying its role as a powerful prior rather than a universal substitute for large-scale task-specific training.

\subsection{Training Efficiency and Practical Considerations}

By freezing the pretrained LLM backbone and adapting it through lightweight LoRA modules, HARLLM drastically reduces the number of trainable parameters. As a result, the entire training process remains computationally efficient and practical. In our experiments, all models were trained on the free tier of Google Colab, using a single commodity GPU, with end-to-end training times consistently below one hour.

This experimental setup corresponds to a realistic resource budget for academic prototyping and small-scale industrial evaluation, rather than specialized high-performance computing environments. The ability to train and fine-tune HARLLM under such constraints highlights its suitability for pervasive and embedded scenarios, where access to large computational resources is often limited. Overall, these results demonstrate that competitive HAR performance can be achieved with pretrained LLMs while maintaining low computational and energy costs during training.

\section{Discussion}

Experimental results provide several insights into the role and limitations of LLMs as backbone architectures for HAR. They show that representations learned through large-scale pretraining can be effectively transferred to inertial time-series data, enabling strong temporal modeling even when the downstream task differs from the original language domain. In particular, rapid convergence under limited supervision highlights the value of LLM backbones as a strong initialization, allowing efficient generalization in data-scarce regimes.

A central finding of our study concerns the specific contribution of the LLM component. Across single-dataset learning, data efficiency, and cross-dataset transfer experiments, HARLLM consistently benefits from the ability of pretrained LLMs to capture long-range temporal dependencies. This capability is  evident in the HHAR dataset, where performance saturates rapidly as the amount of labeled data increases and remains robust even under few-shot fine-tuning. These results suggest that LLMs act primarily as high-level temporal reasoners, effectively modeling activity transitions and long-term context once a suitable low-level representation is provided.

At the same time, our results also reveal important limitations of LLM-centric approaches when local invariances are not sufficiently enforced. This effect is most clearly observed on the RealWorld dataset, which exhibits strong variability in sensor placement and orientation. While HARLLM achieves competitive performance on RealWorld, it does not surpass architectures such as HART that explicitly encode placement invariance through specialized convolutional frontends. This gap indicates that, unlike long-range temporal structure, sensor-placement invariance does not emerge naturally from LLM pretraining and must be addressed at the representation level prior to LLM processing.

The contrast between HHAR and RealWorld highlights two distinct sources of variability in pervasive HAR. HHAR mainly stresses heterogeneity across devices and users while preserving relatively stable sensor placement, making it well suited for LLM-based temporal abstraction and transfer. In contrast, RealWorld introduces strong geometric variability at the signal level, which challenges models lacking explicit inductive biases for orientation and placement invariance; in this case, the limitation of HARLLM stems from the convolutional frontend rather than the LLM itself.

Importantly, this observation clarifies the respective roles of the convolutional encoder and the LLM. The convolutional component is responsible for extracting locally invariant and physically meaningful representations from raw sensor signals, while the LLM operates on these representations to model long-range temporal dependencies and global activity structure. The RealWorld results suggest that improving the convolutional frontend---for instance by incorporating stronger residual or invariant-aware designs---is a more promising direction than increasing LLM capacity. This decomposition reinforces the view that LLMs should not replace domain-specific inductive biases, but rather complement them.

From a broader perspective, these findings support a principled design strategy for LLM-based HAR systems. Rather than treating LLMs as end-to-end signal encoders, our results indicate that their strengths lie in temporal reasoning, transfer learning, and few-shot adaptation. When combined with appropriate signal-level encoders, LLMs provide a flexible and data-efficient backbone that can be adapted to a wide range of pervasive sensing scenarios. Conversely, when local invariances dominate the difficulty of the task, architectural choices at the convolutional level remain critical.

\section{Conclusion}

This paper introduced an alignment-based framework that rethinks how Transformer architectures are used for HAR. Rather than designing task-specific Transformers from scratch, as is common in the HAR literature, we demonstrate that large pretrained language models can be effectively reused as generic temporal backbones, provided that appropriate signal-level preprocessing and alignment mechanisms are applied. In particular, a structured convolutional frontend plays a critical role in bridging the modality gap between inertial time series and language models, enabling LLMs to operate on meaningful and stable temporal representations.

This shift from domain-specific Transformer design to the reuse of existing foundation models has several important implications. By freezing the LLM backbone and adapting it with lightweight LoRA modules, the proposed approach drastically reduces training cost and computational requirements, enabling rapid training on modest hardware. At the same time, pretrained LLMs provide a strong representational prior that supports data-efficient learning, few-shot fine-tuning, and robust cross-dataset transfer, while achieving competitive or state-of-the-art recognition performance without extensive architecture engineering.

Beyond performance, our results clarify the complementary roles of signal processing and large-scale pretraining in HAR systems. While LLMs excel at modeling long-range temporal dependencies and supporting transfer across domains, local invariances related to sensor placement and orientation must be enforced at the convolutional level. This work points toward a more sustainable and scalable paradigm, in which pretrained foundation models are reused across domains to enable frugal, adaptable, and high-quality time-series modeling.

\bibliographystyle{IEEEtran}
\bibliography{sample-base}

@misc{zhang2024largelanguagemodelstime,
      title={Large Language Models for Time Series: A Survey}, 
      author={Xiyuan Zhang and Ranak Roy Chowdhury and Rajesh K. Gupta and Jingbo Shang},
      year={2024},
      eprint={2402.01801},
      archivePrefix={arXiv},
      primaryClass={cs.LG},
      url={https://arxiv.org/abs/2402.01801}, 
}

@inproceedings{ordonez2016deepconv,
  title={Deep convolutional and LSTM recurrent neural networks for multimodal wearable activity recognition},
  author={Ord{\'o}{\~n}ez, Francisco J. and Roggen, Daniel},
  booktitle={Sensors},
  year={2016}
}

@inproceedings{hammerla2016deep,
  title={Deep, convolutional, and recurrent models for human activity recognition using wearables},
  author={Hammerla, Nils Y. and Halloran, Shane and Pl{\"o}tz, Thomas},
  booktitle={IJCAI},
  year={2016}
}

@article{dirgova2021transformer,
  title={Transformer-based human activity recognition},
  author={Dirgov{\'a} Lupt{\'a}kov{\'a}, Ivana et al.},
  journal={Sensors},
  year={2021}
}

@article{ek2022hart,
  title={Transformer-based models to deal with heterogeneous environments in human activity recognition},
  author={Ek, Sannara and Portet, Fran{\c{c}}ois and Lalanda, Philippe},
  journal={Personal and Ubiquitous Computing},
  year={2023}
}

@inproceedings{houlsby2019parameter,
  title={Parameter-Efficient Transfer Learning for {NLP}},
  author={Houlsby, Neil and Giurgiu, Andrei and Jastrzkebski, Stanislaw and Morrone, Bruna and de Laroussilhe, Quentin and Gesmundo, Andrea and Attariyan, Mona and Gelly, Sylvain},
  booktitle={International Conference on Machine Learning},
  pages={2790--2799},
  year={2019},
  organization={PMLR}
}

@inproceedings{li2021prefix,
  title={Prefix-Tuning: Optimizing Continuous Prompts for Generation},
  author={Li, Xiang Lisa and Liang, Percy},
  booktitle={Proceedings of the 59th Annual Meeting of the Association for Computational Linguistics and the 11th International Joint Conference on Natural Language Processing (Volume 1: Long Papers)},
  pages={4582--4597},
  year={2021}
}

@inproceedings{lester2021power,
  title={The Power of Scale for Parameter-Efficient Prompt Tuning},
  author={Lester, Brian and Al-Rfou, Rami and Constant, Noah},
  booktitle={Proceedings of the 2021 Conference on Empirical Methods in Natural Language Processing},
  pages={3045--3059},
  year={2021}
}

@article{hu2021lora,
  title={Lo{RA}: Low-Rank Adaptation of Large Language Models},
  author={Hu, Edward J. and Shen, Yelong and Wallis, Phillip and Allen-Zhu, Zeyuan and Li, Yuanzhi and Wang, Shean and Wang, Lu and Chen, Weizhu},
  journal={arXiv preprint arXiv:2106.09685},
  year={2021}
}

@article{radford2019language,
  title={Language Models are Unsupervised Multitask Learners},
  author={Radford, Alec and Wu, Jeffrey and Child, Rewon and Luan, David and Amodei, Dario and Sutskever, Ilya},
  journal={OpenAI Blog},
  year={2019}
}

@article{touvron2023llama,
  title={{LLaMA}: Open and Efficient Foundation Language Models},
  author={Touvron, Hugo and Lavril, Thibaut and Izacard, Gautier and Martinet, Xavier and Lachaux, Marie-Anne and Lacroix, Timoth{\'e}e and Rozi{\`e}re, Baptiste and Goyal, Naman and Hambro, Eric and Azhar, Faisal and others},
  journal={arXiv preprint arXiv:2302.13971},
  year={2023}
}

@article{bai2023qwen,
  title={Qwen Technical Report},
  author={Bai, Jinze and Bai, Shuai and Yang, Shusheng and Wang, Shijie and Tan, Sinan and Wang, Peng and Lin, Junyang and Zhou, Chang and Zhou, Jingren},
  journal={arXiv preprint arXiv:2309.16609},
  year={2023}
}

@article{ignatov2018real,
  title={Real-time human activity recognition from accelerometer data using Convolutional Neural Networks},
  author={Ignatov, Andrey},
  journal={Applied Soft Computing},
  volume={62},
  year={2018},
  publisher={Elsevier}
}

@article{IFConvTransformer2022,
author = {Zhang, Ye and Wang, Longguang and Chen, Huiling and Tian, Aosheng and Zhou, Shilin and Guo, Yulan},
title = {IF-ConvTransformer: A Framework for Human Activity Recognition Using IMU Fusion and ConvTransformer},
year = {2022},
issue_date = {July 2022},
journal = {Proc. ACM Interact. Mob. Wearable Ubiquitous Technol.},
address = {New York, NY, USA},
volume = {6},
number = {2}
}

@INPROCEEDINGS{10207558,
  author={Presotto, Riccardo and Ek, Sannara and Civitarese, Gabriele and Portet, François and Lalanda, Philippe and Bettini, Claudio},
  booktitle={2023 IEEE International Conference on Smart Computing (SMARTCOMP)}, 
  title={Combining Public Human Activity Recognition Datasets to Mitigate Labeled Data Scarcity}, 
  year={2023},
  volume={},
  number={},
  pages={33-40},
  doi={10.1109/SMARTCOMP58114.2023.00022}}

@article{ordonez2016deep,
  title={Deep convolutional and lstm recurrent neural networks for multimodal wearable activity recognition},
  author={Ord{\'o}{\~n}ez, Francisco Javier and Roggen, Daniel},
  journal={Sensors},
  volume={16},
  number={1},
  pages={115},
  year={2016},
  publisher={Multidisciplinary Digital Publishing Institute}
}

@inproceedings{
dosovitskiy2021an,
title={An Image is Worth 16x16 Words: Transformers for Image Recognition at Scale},
author={Alexey Dosovitskiy and Lucas Beyer and Alexander Kolesnikov and Dirk Weissenborn and Xiaohua Zhai and Thomas Unterthiner and Mostafa Dehghani and Matthias Minderer and Georg Heigold and Sylvain Gelly and Jakob Uszkoreit and Neil Houlsby},
booktitle={International Conference on Learning Representations},
year={2021},
}

@inproceedings{Anguita2013APD,
  author    = {Davide Anguita and
               Alessandro Ghio and
               Luca Oneto and
               Xavier Parra and
               Jorge Luis Reyes{-}Ortiz},
  title     = {A Public Domain Dataset for Human Activity Recognition using Smartphones},
  booktitle = {21st European Symposium on Artificial Neural Networks, {ESANN} 2013,
               Bruges, Belgium, April 24-26, 2013},
  year      = {2013},
}

@INPROCEEDINGS{realword,
  author={T. {Sztyler} and H. {Stuckenschmidt}},
  booktitle={2016 IEEE International Conference on Pervasive Computing and Communications (PerCom)}, 
  title={On-body localization of wearable devices: An investigation of position-aware activity recognition}, 
  year={2016},
  volume={},
  number={},
  pages={1-9},}

@article{Ignatov2018RealtimeHA,
  title={Real-time human activity recognition from accelerometer data using Convolutional Neural Networks},
  author={Andrey D. Ignatov},
  journal={Appl. Soft Comput.},
  year={2018}
}

@article{wang2019deep,
  title={Deep learning for sensor-based activity recognition: A survey},
  author={Wang, Jindong and Chen, Yiqiang and Hao, Shuji and Peng, Xiaohui and Hu, Lisha},
  journal={Pattern recognition letters},
  volume={119},
  pages={3--11},
  year={2019},
  publisher={Elsevier}
}

@article{hochreiter1997long,
  title={Long short-term memory},
  author={Hochreiter, Sepp and Schmidhuber, J{\"u}rgen},
  journal={Neural computation},
  volume={9},
  number={8},
  pages={1735--1780},
  year={1997},
  publisher={MIT press}
}

@article{onofri2016survey,
  title={A survey on using domain and contextual knowledge for human activity recognition in video streams},
  author={Onofri, Leonardo and Soda, Paolo and Pechenizkiy, Mykola and Iannello, Giulio},
  journal={Expert Systems with Applications},
  volume={63},
  pages={97--111},
  year={2016},
  publisher={Elsevier}
}

@article{vaswani2017attention,
  title={Attention is all you need},
  author={Vaswani, Ashish and Shazeer, Noam and Parmar, Niki and Uszkoreit, Jakob and Jones, Llion and Gomez, Aidan N and Kaiser, {\L}ukasz and Polosukhin, Illia},
  journal={Advances in neural information processing systems},
  volume={30},
  year={2017}
}

@article{ek2024comparing,
  title={Comparing Self-Supervised Learning Techniques for Wearable Human Activity Recognition},
  author={Ek, Sannara and Presotto, Riccardo and Civitarese, Gabriele and Portet, Fran{\c{c}}ois and Lalanda, Philippe and Bettini, Claudio},
  journal={arXiv preprint arXiv:2404.15331},
  year={2024}
}

@inproceedings{stisen2015smart,
  title={Smart devices are different: Assessing and mitigatingmobile sensing heterogeneities for activity recognition},
  author={Stisen, Allan and Blunck, Henrik and Bhattacharya, Sourav and Prentow, Thor Siiger and Kj{\ae}rgaard, Mikkel Baun and Dey, Anind and Sonne, Tobias and Jensen, Mads M{\o}ller},
  booktitle={Proceedings of the 13th ACM conference on embedded networked sensor systems},
  pages={127--140},
  year={2015}
}

\end{document}